%% file: main.tex
\pdfoutput=1

\documentclass[11pt]{article}

\usepackage[final]{acl}

\usepackage{times}
\usepackage{latexsym}

\usepackage[T1]{fontenc}

\usepackage[utf8]{inputenc}

\usepackage{microtype}

\usepackage{inconsolata}

\usepackage{graphicx}
\usepackage[table]{xcolor}
\usepackage{adjustbox}
\usepackage{enumitem}
\usepackage{booktabs}

\usepackage{colortbl}
\usepackage{geometry}
\usepackage{adjustbox}
\usepackage{makecell}
\usepackage{multirow}

\definecolor{green25}{RGB}{200, 255, 200}
\definecolor{red20}{RGB}{255, 200, 200}

\usepackage{amsmath}   %
\usepackage{amssymb}   %

\title{Mind the Gap: Benchmarking LLM Uncertainty and Calibration with Specialty-Aware Clinical QA and Reasoning-Based Behavioural Features}

\author{
  \textbf{Alberto Testoni\textsuperscript{1,2}},
  \textbf{Iacer Calixto\textsuperscript{1,2}}
\\
\\
  \textsuperscript{1}Department of Medical Informatics, Amsterdam University Medical Center,\\
                     University of Amsterdam, Amsterdam, The Netherlands,\\
  \textsuperscript{2}Amsterdam Public Health, Methodology, Amsterdam, The Netherlands,
\\
  \small{
    \textbf{Correspondence:} \href{mailto:a.testoni@amsterdamumc.nl}{a.testoni@amsterdamumc.nl}; \href{mailto:i.coimbra@amsterdamumc.nl}{i.coimbra@amsterdamumc.nl}
  }
}

\begin{document}
\maketitle
\begin{abstract}

Reliable uncertainty quantification (UQ) is essential when employing large language models (LLMs) in high-risk domains such as clinical question answering (QA). In this work, we evaluate uncertainty estimation methods for clinical QA focusing, for the first time, on eleven clinical specialties and six question types, and across ten open-source LLMs (general-purpose, biomedical, and reasoning models), alongside representative proprietary models. We analyze score-based UQ methods, present a case study introducing a novel lightweight method based on behavioral features derived from reasoning-oriented models, and examine conformal prediction as a complementary set-based approach. Our findings reveal that uncertainty reliability is not a monolithic property, but one that depends on clinical specialty and question type due to shifts in calibration and discrimination. Our results highlight the need to select or ensemble models based on their distinct, complementary strengths and clinical use.

\end{abstract}

\input{parts/1-new-intro-iacer}
\input{parts/2-related}

\input{parts/3-datasets}
\input{parts/3-methods}

\input{parts/4-results}

\input{parts/5-CF}

\input{parts/6-conclusions}

\section*{Limitations}

In the following, we discuss some limitations of our experimental setup.

\paragraph{Open-ended generation mapped to MCQA.} While we benchmark on multiple-choice question answering for comparability and ground truth clarity, our setup involves open-ended generation: models first produce free-form answers and reasoning traces, which we then map to the multiple-choice options for evaluation.  We believe that fully open-ended QA is crucial for clinical realism, but it comes with significant challenges in defining gold standards. We believe this focus is justified as a first step: to the best of our knowledge, ours is the first study to systematically evaluate UQ across model types, medical specialties, and question types at this scale. Nonetheless, future work should extend this analysis to more open-ended clinical scenarios.

\paragraph{Automated prediction and annotation pipelines.} We rely on automated extraction of model predictions and question-type annotations. Although our heuristics and model-assisted annotations are validated against manual labels with high agreement, some residual noise may persist, especially for ambiguous questions or borderline categories (e.g., ``Other'').

\paragraph{Behavioral features as exploratory proxies.} We use behavioral features as proxies for uncertainty in an exploratory case study, as highlighted throughout the main text (Section \ref{subsec:behavioural}). Validation with domain experts or clinicians is planned as an important next step to confirm their practical value. While we show that behavioral features offer a promising, interpretable alternative to multi-sample methods like Semantic Entropy, our feature space is deliberately simple and task-specific. Generalizing these findings to other tasks, languages, or reasoning paradigms will require broader validation and potentially more sophisticated feature design.

\paragraph{Sources of uncertainty.} We do not investigate in detail sources of uncertainty. We believe future work should look into promising approaches in this direction, such as \citet{yang2025maqa}.

\paragraph{Post-hoc calibration.} We do not apply post-hoc calibration techniques such as temperature scaling. Our focus is on comparing raw uncertainty estimates without additional tuning. Applying temperature scaling consistently is challenging in our setup, as models differ in tokenization schemes and output structure  -- e.g., in how they format answers (direct label vs. more complex sentences), affecting how likelihoods are computed and making calibration non-trivial. We leave the integration of model-specific calibration as an important direction for future work. As for additional considerations on the temperature setup and decoding details, see Appendix~\ref{appendix:decoding}.

\paragraph{Uncertainty communication.} We benchmark uncertainty estimates (including confidence elicitation) but do not study how uncertainty should be communicated to end users \cite{ulmer2025anthropomimetic}; verbalized uncertainty is shaped by contextual and sociopragmatic factors and can be biased by alignment for helpfulness, raising additional challenges for clinician-facing deployment.

\paragraph{Coverage for ChatGPT models.} Our analysis of OpenAI ChatGPT models (see Appendix \ref{appendix:gpt}) is limited to Semantic Entropy, which proved to be the most robust UQ method. Future work should evaluate other methods, such as Confidence Elicitation and Conformal Prediction, also for these models.

\section*{Acknowledgments}
This publication is part of the project CaRe-NLP with file number NGF.1607.22.014 of the research programme AiNed Fellowship Grants which is (partly) financed by the Dutch Research Council (NWO). We thank the members of the NLP4Health Lab and KIK AI at Amsterdam UMC, as well as the DMG group at the ILLC, University of Amsterdam, for their insightful feedback and helpful suggestions at different stages of this work.

\bibliography{anthology,custom}

\appendix

\section*{Appendix}
\label{sec:appendix}

\section{Model Details and Licenses}
\label{appendix:models}
The \texttt{LLaMA} family of models and \texttt{LLaMA3-OpenBioLLM-70B} are distributed under the Meta AI community license agreement. \texttt{Mistral-7B-v0.2}, \texttt{Mixtral-8x7B-v0.1}, \texttt{s1.1-32B} are released under Apache license 2.0. \texttt{DeepSeek-R1-Distill-Llama-70B} is distributed under MIT license. \texttt{Bio-Medical-Llama-3-8B} is released under Bio-Medical-Llama-3-8B LLM License (Non-Commercial Use Only). All models are accessed through Hugging Face (\url{https://huggingface.co/}). We run inference on a cluster with 4 A100 GPUs and a batch size of 8. We run small and medium models (3B-32B parameters) on 2 GPUs and larger models on 4 GPUs. We use the \texttt{vLLM} library. Running inference takes on average a couple of hours for small models and 24 hours for larger models and reasoning models, which produce longer model responses. We used all datasets in accordance with their licenses and consented use policies.

\begin{table*}[t]
\centering
\small
\begin{tabular}{lcccc}
\toprule
\textbf{Model} & \textbf{Accuracy} & \textbf{AUROC} & \textbf{ECE} & \textbf{Brier Score} \\
\midrule
Llama-3.1-8B-Instruct & 0.583 & 0.806 & 0.048 & 0.176 \\
Llama-3.1-70B-Instruct & 0.784 & 0.873 & 0.040 & 0.104 \\
Llama-3.2-3B-Instruct & 0.422 & 0.868 & 0.101 & 0.153 \\
Llama-3.3-70B-Instruct & 0.774 & 0.858 & 0.036 & 0.112 \\
Mistral-7B-Instruct-v0.2 & 0.474 & 0.665 & 0.078 & 0.293 \\
Mixtral-8x7B-Instruct-v0.1 & 0.602 & 0.708 & 0.040 & 0.225 \\
DeepSeek-R1-70B & 0.767 & 0.898 & 0.119 & 0.113 \\
s1.1-32B & 0.837 & 0.852 & 0.081 & 0.090 \\
Llama3-OpenBioLLM-70B & 0.699 & 0.856 & 0.093 & 0.140 \\
Bio-Medical-Llama-3-8B & 0.666 & 0.781 & 0.030 & 0.169 \\
\bottomrule
\end{tabular}
\caption{Evaluation results with $T=1$ (uncertainty estimates extracted as Semantic Entropy).}
\label{tab:model_eval_T1}
\end{table*}

\section{Prompts}
\label{appendix:prompts}
\paragraph{Model Responses}
In order to get the model responses to the input question, we employ the following prompt (example of an instantiation in parenthesis): 

\begin{quote}
\small
Question: (A 59-year-old overweight woman presents to the urgent care clinic with the complaint of severe abdominal pain for the past 2 hours. She also complains of a dull pain in her back with nausea and vomiting several times. Her pain has no relation with food. Her past medical history is significant for recurrent abdominal pain due to cholelithiasis. Her father died at the age of 60 with some form of abdominal cancer. Her temperature is 37°C (98.6°F), respirations are 15/min, pulse is 67/min, and blood pressure is 122/98 mm Hg. Physical exam is unremarkable. However, a CT scan of the abdomen shows a calcified mass near her gallbladder. Which of the following diagnoses should be excluded first in this patient?)

Provide a concise answer and start your response with the letter of the selected option in square brackets. Options:

(

[A] Acute cholecystitis

[B] Gallbladder cancer

[C] Choledocholithiasis

[D] Pancreatitis"

)

Your Answer: 
\end{quote}

For models that come with a recommended system prompt, we use it as provided by the model developers/authors.

\paragraph{Mapping Responses to the Corresponding Option}
We use the following prompt to map model responses that do not follow the required format (i.e., the letter of the selected option in square brackets) to the corresponding answer choice. Before running the annotation, we inspected a random sample of 100 ill-formatted responses and found no need to include an ``other''/``invalid'' class for this task. Placeholders in  curly brackets.

\begin{quote}
\small
You will be given a multiple-choice question, four answer options labeled A, B, C, and D, and a free-form response.

Question:
{question}

Options:

A. \{option a\}

B. \{option b\}

C. \{option c\}

D. \{option d\}

Response:
\{response\}

Your task is to determine which of the four options the response refers to. Output only the corresponding letter in square brackets, like [X]. Do not include any explanation or additional text.

Your annotation must be exactly one of the following: [A], [B], [C], or [D].

If the response includes a different letter, you must still map it to one of the four valid options above, based on the content of the response.

Keep in mind that the response may be poorly formatted or contain irrelevant letters—focus only on identifying the most likely intended option.

If multiple options seem plausible, choose the one that is most strongly implied by the response alone, without relying on external knowledge or context.

Your annotation:

\end{quote}

\paragraph{Question Type Annotation}
We used the following prompt to assign each question to a specific question type (placeholders in curly brackets):

\begin{quote}
\small
You are a medical expert assistant. Your task is to classify the type of each medical question according to the following categories:

1. Diagnosis – Identifying a disease or condition.

2. Treatment – Interventions like medication, surgery, or therapy.

3. Diagnostic Test – Lab tests, imaging, or diagnostic procedures.

4. Definition – Asking for an explanation of a medical concept.

5. Procedure/Operation – Questions about medical or surgical procedures.

6. Other

Assign the most appropriate type based on the question intent. 

Classify the following question. 
Provide only the corresponding label number (1–6) in square brackets as your response.

Q: \{\}

Question type:

\end{quote}

\section{Decoding Details}
\label{appendix:decoding}
For experiments requiring multiple samples, we reviewed commonly suggested decoding configurations for nucleus sampling to approximate how language models are typically used in practice. Based on this review, we set the top-\(p\) value to 0.9 and the temperature to 0.6. As an exploratory analysis, we also repeated our experiments using a higher temperature of 1.0 (Table~\ref{tab:model_eval_T1}). We found that this change had little effect on the overall ranking of models in terms of accuracy or on the calibration and discrimination performance of uncertainty estimates derived from Semantic Entropy. However, for \texttt{DeepSeek-R1-70B}, we observed a decline in both accuracy and calibration. This effect warrants further investigation, which would require a held-out validation set and a systematic analysis across a broader range of temperature values. Such an analysis might ultimately support model-specific or even task-specific decoding configurations, varying across models, medical specialties, or question types. In this work, however, we adhere to decoding settings—temperature and top-\(p\)—that reflect typical real-world usage, especially for users with limited technical knowledge. As for annotating question types and mapping the responses to the corresponding option, we employ greedy decoding.  

\begin{figure*}[t]
	\centering
	\includegraphics[width=0.85\linewidth]{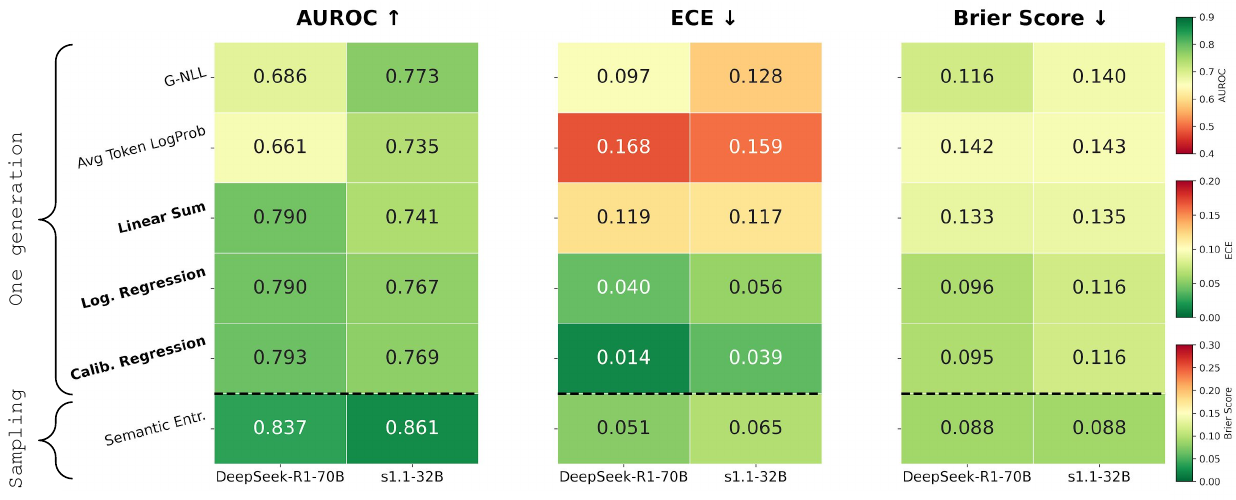}
	\caption{Our proposed methods, based on behavioral features (shown in bold), offer strong performance across AUROC, ECE, and Brier Score, approaching that of Semantic Entropy, which instead relies on multiple generations via sampling.
	}
	\label{fig:ling_features_smedqa_appendix}
\end{figure*}

\begin{figure*}[t]
	\centering
	\includegraphics[width=0.9\linewidth]{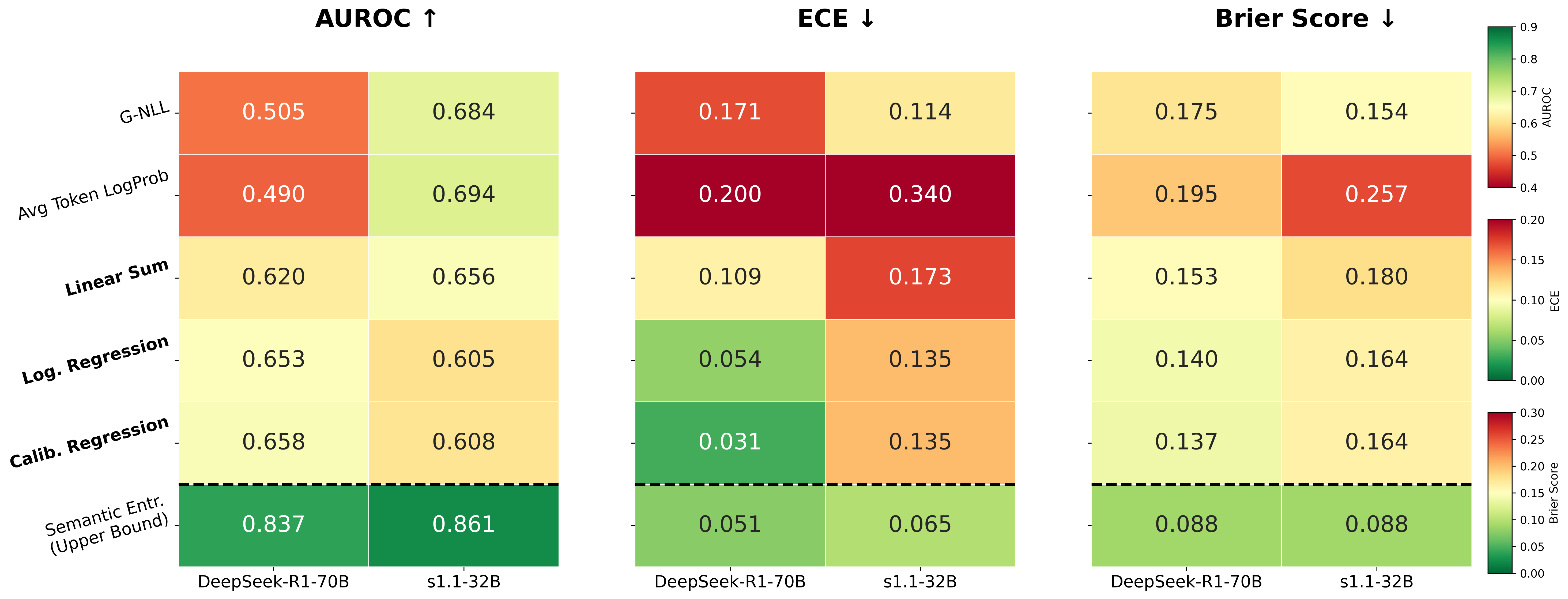}
	\caption{Comparison of UQ methods for \texttt{DeepSeek-R1-70B} and \texttt{s1.1-32B} for uncommon medical specialties.
}
	\label{fig:ling_features_medexqa}
\end{figure*}

\section{Behavioral Features in Reasoning Traces}
\label{appendix:behavioural}

In this section, we provide additional details on the experiments using behavioral features for reasoning models (Section \ref{subsec:behavioural}), including regression-based results. In earlier stages of the project, we also tested alternative approaches such as Random Forests, a simple MLP, and sentence embeddings from \texttt{SentenceTransformer}, all of which performed worse than the models reported here and were therefore discarded. We further experimented with subsets of the three behavioral features, but observed consistent performance drops.

\subsection{Additional details on the method}
Given the features described in Section \ref{subsec:behavioural}, the first is a simple normalized sum of the three features, referred to as the \textit{Linear Sum} method. Let \( x_1 \), \( x_2 \), and \( x_3 \) denote the normalized values of the three features, and compute the uncertainty estimate as \( u = x_1 + x_2 + x_3 \). The second approach treats the three features as input to a supervised model: either a \textit{Logistic Regression} or a \textit{Calibration Regression} (the former models a sigmoid likelihood, the latter fits a linear function with L2 regularization as \textit{Ridge Regression}). \textit{Calibration Regression} is the method discussed in Section \ref{subsec:behavioural}, while here we provide details on all methods. To train these models, we use ground-truth correctness labels (1 for correct, 0 for incorrect) associated with each response. Formally, the model learns a function \( f: \mathbb{R}^3 \rightarrow [0,1] \). Although the regression models are trained to predict correctness, we hypothesize that their outputs can serve as proxies for confidence. We define uncertainty as their complement, \( u = 1 - f(\mathbf{x}) \), enabling direct comparison with other uncertainty estimation methods.
To ensure a fair comparison and avoid overfitting, we train the regression models on 40\% of the dataset while evaluating on the remaining 60\%, preserving the original distribution of medical specialties. All responses are generated via greedy decoding to enable direct comparison with single-pass baselines such as G-NLL and average token probability. We assess the quality of the behavioral uncertainty signals using AUROC, ECE, and Brier Score. To further disentangle model behavior, we report results separately for common and uncommon medical specialties. We include results with Semantic Entropy for comparison, noting that it relies on repeated sampling to estimate uncertainty.

The results, reported in Figure \ref{fig:ling_features_smedqa_appendix}, reveal that our lightweight behavioral feature-based methods—\textit{Linear Sum}, \textit{Logistic Regression}, and \textit{Calibration Regression}— outperform other sentence-level methods and closely match the performance of Semantic Entropy, while requiring only a single forward pass per input. This pattern holds for both reasoning LLMs, \texttt{DeepSeek-R1-70B} and \texttt{s1.1-32B}. Even the simple \textit{Linear Sum} method outperforms token-level baselines like G-NLL and average token log-probability in most cases, particularly in discrimination (AUROC). \textit{Calibration Regression} performs consistently well, especially in terms of calibration. In Section \ref{appendix:regression_coefficients}, we provide further details on the regression model coefficients. Results on less-represented specialties are reported in Figure \ref{fig:ling_features_medexqa}. They confirm the effectiveness of our approach for \texttt{DeepSeek-R1-70B}, while showing limited improvement for \texttt{s1.1-32B}, highlighting the need for model-specific tuning or more expressive behavioral features in lower-resource settings. Our case study offers a starting point for future work that should validate this approach across a broader range of tasks, datasets, and models. We view our contribution as an initial step toward establishing a more general framework for behavioral uncertainty estimation that leverages the output of recently released reasoning LLMs.

\subsection{Example of Model Response}
\label{appendix:ex_model_behavioral}

In the sample model response below, generated by \texttt{DeepSeek-R1-70B}, we highlight in bold some of the behavioural features examined in our work, including self-verification questions and ``wait'' tokens (we highlight relevant sentences below). 

\begin{quote}
\small 

\textbf{Question}: A 28-year-old male presents to the emergency department with chest pain. He reports that one hour ago he was climbing the stairs in his home when he suddenly developed excruciating chest pain. He describes the pain as ripping and radiating to the back. The patient’s past medical history is significant for a heart murmur that is being followed by his outpatient cardiologist. He takes no medications. The patient works as a carpenter and reports being monogamous with his wife. He denies any history of intravenous drug use or recent travel. In the emergency department, the patient’s temperature is 99.4°F (37.4°C), blood pressure is 94/56 mmHg, pulse is 121/min, and respirations are 14/min. On physical exam, the patient appears to be in moderate distress and is clutching his chest. He is tall and has joint hypermobility. His breath sounds are equal and full bilaterally and has weak distal pulses.

Which of the following is most likely to be found in this patient?

[A] Late systolic crescendo murmur at the apex with mid-systolic click

[B] Mid-diastolic murmur at the left lower sternal border

[C] Systolic ejection murmur at the right second intercostal space

[D] Holosystolic, high-pitched blowing murmur at the lower left sternal border

\textbf{Model Response}: 

[C]

<think>

Okay, so I'm trying to figure out this medical question. Let me read through it again and break it down step by step.

The patient is a 28-year-old male presenting with chest pain. He was climbing stairs when he suddenly developed excruciating, ripping chest pain that radiates to the back. [...]

\textbf{Wait, but aortic regurgitation typically presents with a diastolic murmur, not systolic. So maybe I'm mixing things up. }Alternatively, if the patient has a bicuspid aortic valve, which is common in Marfan, that could cause a systolic ejection murmur. Also, in aortic dissection, the murmur might not be present unless there's associated aortic regurgitation.

D. Holosystolic, high-pitched blowing murmur at the lower left sternal border: This is typical of mitral regurgitation. The holosystolic nature suggests blood flowing back through the mitral valve throughout systole. But again, the patient's presentation points more towards aortic issues.

\textbf{Wait, but the patient's chest pain and physical exam suggest aortic dissection. So, what murmur is associated with that?} If the dissection leads to aortic regurgitation, the murmur would be a diastolic one, like an early diastolic murmur best heard at the left sternal border. But none of the options are diastolic.

Alternatively, if the patient has a bicuspid aortic valve, which is a common congenital heart defect, that could present with a systolic ejection murmur at the right second intercostal space. Bicuspid aortic valve is associated with an increased risk of aortic dissection, especially in the context of Marfan syndrome or other connective tissue disorders. [...]

\textbf{Wait, but earlier I thought that aortic regurgitation would present with a diastolic murmur. So, if the dissection caused aortic regurgitation, the murmur would be diastolic, but that's not one of the options. So perhaps the murmur is from the underlying bicuspid valve, which is a systolic ejection murmur.}

Yes, that makes sense. The patient's heart murmur is likely from the bicuspid aortic valve, which is a systolic ejection murmur heard at the right second intercostal space. So the answer would be option C.
</think>

The patient's presentation of sudden, severe, tearing chest pain radiating to the back, along with tall stature, joint hypermobility, and weak distal pulses, strongly suggests aortic dissection. This condition is often associated with underlying bicuspid aortic valve, which typically presents with a systolic ejection murmur at the right second intercostal space. 

[C] Systolic ejection murmur at the right second intercostal space

\end{quote}

\subsection{Regression Coefficients}
\label{appendix:regression_coefficients}

Table \ref{tab:regression_coeffs} reports the regression coefficients for the methods based on behavioural features of reasoning models. For both models, a higher token count is associated with lower accuracy and worse calibration, suggesting that verbosity reflects uncertainty or confusion. This effect is more pronounced in \texttt{s1.1-32B}, indicating it may overgenerate even when less confident.

\begin{table}[t]
\centering
\small
\begin{tabular}{lrrrr}
\hline
 & \multicolumn{2}{c}{\textbf{Logistic Regr.}} & \multicolumn{2}{c}{\textbf{Calibration Regr.}} \\
\textbf{Feature} & \textbf{DeepSeek} & \textbf{s1.1} & \textbf{DeepSeek} & \textbf{s1.1} \\
\hline
tokens          & -0.6476 & -1.4611 & -0.0925 & -0.1527 \\
questions       & -0.1052 &  0.0755 & -0.0196 & -0.0181 \\
waits           & -0.3456 &  0.3961 & -0.0839 &  0.0708 \\
\textit{bias}&  1.8559 &  1.9407 &  0.8269 &  0.8556 \\
\hline
\end{tabular}
\caption{Regression Coefficients.}
\label{tab:regression_coeffs}
\end{table}

\begin{figure}[t]
	\centering
	\includegraphics[width=0.9\linewidth]{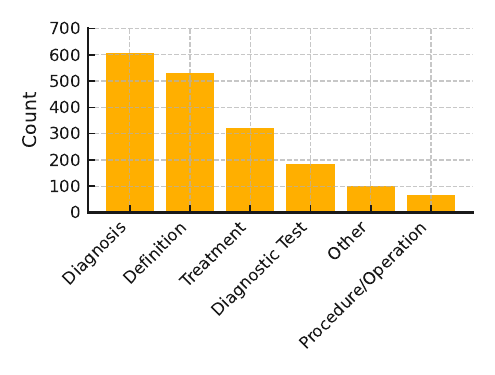}
	\caption{Frequency of question types.
}
	\label{fig:qtype_freq}
\end{figure}

\section{Additional Details on Uncertainty Quantification Methods}

First, we chose not to include confidence elicitation methods that directly prompt LLMs to report a confidence score. While such approaches have been explored in prior work, their reliability remains limited. In particular, \citet{savage2025large} report poor performance and calibration when relying on model-generated confidence scores.

Second, for methods based on log-probabilities (logprobs), we carefully considered how to handle outputs from models that include explicit reasoning traces. A notable example is \texttt{DeepSeek-R1-70B}, whose responses are structured with clearly delimited reasoning segments, marked by special tokens such as \texttt{<think>}. In these cases, a key decision is whether to compute uncertainty scores over the entire response (including the reasoning trace) or only over the final answer portion. After empirical inspection and preliminary analyses, we found that restricting the computation of logprobs to the final answer (excluding the reasoning trace) produced more reliable and comparable uncertainty estimates. 

\section{Question Type Distribution}
\label{appendix:qtype}

Figure \ref{fig:qtype_freq} reports the frequency of the question types analyzed in our experiments.

\section{GPT Results with Semantic Entropy}
\label{appendix:gpt}

Tables \ref{tab:gpt_4.1}, \ref{tab:gpt_4.1_mini}, and \ref{tab:gpt_4o_mini} report the performance of three representative OpenAI models (\texttt{GPT-4.1}, \texttt{GPT-4.1-mini}, and \texttt{GPT-4o-mini}) using Semantic Entropy to extract uncertainty estimates across common and uncommon medical specialties. These models show strong accuracy and calibration (average ECE and Brier score) but surprisingly poor discrimination (1-AUROC), sometimes even below smaller open-source models (see Figure \ref{fig:plot_discrimination_calibration}). Importantly, we once again observe differences across specialties, reinforcing our conclusions. We observe an effect of model size, but less pronounced than expected. \texttt{GPT-4.1} is generally more accurate and better calibrated than \texttt{GPT-4.1-mini}, its small counterpart. However, this does not hold for all specialties: for instance, for questions in the \textit{Clinical Laboratory Science} domain, the uncertainty estimates extracted from \texttt{GPT-4.1-mini} are better calibrated. In line with the results in Section \ref{sec:results}, GPT models show consistently weak performance in the \textit{Speech Language Pathology} specialty. Taken together, these results suggest that the proprietary models evaluated achieve strong accuracy and calibration, yet fall short in producing uncertainty estimates with sufficient discriminative power. Furthermore, the unavailability of log probabilities and reasoning traces limits transparency and constrains deeper analysis.

\begin{table}[htbp]
\centering
\resizebox{\columnwidth}{!}{%
\begin{tabular}{llll}
\toprule
Specialty & Accuracy ↑ & 1-AUROC ↓ & Avg(ECE,Brier) ↓ \\
\midrule
Cardiology & 0.8945 & 0.1005 & 0.0232 \\
Gastroenterology & 0.9007 & 0.2528 & 0.0223 \\
Infectious diseases & 0.8890 & 0.2430 & 0.0230 \\
Neurology & 0.8752 & 0.2466 & 0.0256 \\
Obs. and gyn. & 0.8986 & 0.2492 & 0.0401 \\
Pediatrics & 0.8986 & 0.1559 & 0.0290 \\
Biomedical Eng. & 0.8853 & 0.1946 & 0.0292 \\
Clinical Lab. Science & 0.9320 & 0.2931 & 0.0135 \\
Clinical Psychology & 0.9387 & 0.0453 & 0.0189 \\
Occupational Therapy & 0.8942 & 0.1176 & 0.0176 \\
Speech-Lang. Path. & 0.3123 & 0.5021 & 0.0159 \\
\bottomrule
\end{tabular}}
\caption{\texttt{GPT-4.1} results.}
\label{tab:gpt_4.1}
\end{table}

\begin{table}[htbp]
\centering
\resizebox{\columnwidth}{!}{%
\begin{tabular}{llll}
\toprule
Specialty & Accuracy ↑ & 1-AUROC ↓ & Avg(ECE,Brier) ↓ \\
\midrule
Cardiology & 0.8593 & 0.2784 & 0.0434 \\
Gastroenterology & 0.8290 & 0.3944 & 0.0283 \\
Infectious diseases & 0.8310 & 0.2688 & 0.0358 \\
Neurology & 0.8276 & 0.2802 & 0.0361 \\
Obs. and gyn. & 0.8379 & 0.2885 & 0.0252 \\
Pediatrics & 0.8572 & 0.1477 & 0.0396 \\
Biomedical Eng. & 0.8706 & 0.2255 & 0.0256 \\
Clinical Lab. Science & 0.9153 & 0.1851 & 0.0184 \\
Clinical Psychology & 0.9377 & 0.1446 & 0.0145 \\
Occupational Therapy & 0.8974 & 0.2672 & 0.0199 \\
Speech-Lang. Path. & 0.3100 & 0.4567 & 0.0133 \\
\bottomrule
\end{tabular}}
\caption{\texttt{GPT-4.1-mini} results.}
\label{tab:gpt_4.1_mini}
\end{table}

\begin{table}[htbp]
\centering
\resizebox{\columnwidth}{!}{%
\begin{tabular}{llll}
\toprule
Specialty & Accuracy ↑ & 1-AUROC ↓ & Avg(ECE,Brier) ↓ \\
\midrule
Cardiology & 0.7800 & 0.2923 & 0.0477 \\
Gastroenterology & 0.7552 & 0.2840 & 0.0406 \\
Infectious diseases & 0.7352 & 0.2900 & 0.0537 \\
Neurology & 0.7131 & 0.3006 & 0.0612 \\
Obs. and gyn. & 0.7324 & 0.3463 & 0.0448 \\
Pediatrics & 0.7807 & 0.1904 & 0.0465 \\
Biomedical Eng. & 0.8531 & 0.2442 & 0.0251 \\
Clinical Lab. Science & 0.8680 & 0.2147 & 0.0279 \\
Clinical Psychology & 0.9151 & 0.1071 & 0.0246 \\
Occupational Therapy & 0.8852 & 0.2529 & 0.0265 \\
Speech-Lang. Path. & 0.3085 & 0.4826 & 0.0223 \\
\bottomrule
\end{tabular}}
\caption{\texttt{GPT-4o-mini} results.}
\label{tab:gpt_4o_mini}
\end{table}

\section{Conformal Prediction for Multiple-Choice QA}
\label{appendix:cp}

\subsection{Method}
We adopt a conformal prediction framework for multiple-choice question answering (MCQA), following the methodology proposed by \citet{ke2025correctness} and summarized below.

\subsubsection*{Step 1: Calibration}

\begin{enumerate}
    \item[\textbf{1A.}] \textbf{Sample Generations.} For each calibration question with known correct answer, sample $M$ model outputs and extract the selected answer option.
    
    \textit{Example:} \\
    \textbf{Question:} Which organ produces insulin? \\
    \textbf{Correct answer:} B \\
    \textbf{Sampled answers (M=20):} B, B, C, B, A, B, B, B, B, C, B, B, B, B, B, B, B, B, B, B
    
    \textbf{Empirical frequencies:} A: 1, B: 17, C: 2, D: 0 \\
    $\Rightarrow \hat{p}_{\text{correct}} = \frac{17}{20} = 0.85$, so $S = 1 - 0.85 = 0.15$
    
    \item[\textbf{1B.}] \textbf{Compute Non-conformity Scores.} Repeat the above process for all calibration questions to compute a set of scores $S_1, S_2, \dots, S_N$.

    \item[\textbf{1C.}] \textbf{Estimate the Threshold.} Sort the $S$ values and set $\hat{q}$ as the $(1 - \alpha)$ quantile. 
    
    \textit{Example:} \\
    Calibration scores: [0.05, 0.10, 0.15, 0.20, ..., 0.85] \\
    For $\alpha = 0.1$, 90th percentile $\Rightarrow \hat{q} = 0.65$
\end{enumerate}

\noindent In our setup, we extract 320 questions per specialty from the S-MedQA validation set to serve as calibration set.

\subsubsection*{Step 2: Prediction}
\begin{enumerate}
    \item[\textbf{2A.}] \textbf{Sample Generations.} For a test-time question, sample $M$ model outputs and extract the selected options.

    \textit{Example:} \\
    \textbf{Question:} What is the most common cause of X? \\
    \textbf{Sampled answers (M=20):} B, B, B, C, B, B, B, A, B, B, B, B, B, B, B, B, B, B, B, B \\
    \textbf{Frequencies:} A: 1, B: 18, C: 1, D: 0
    
    \item[\textbf{2B.}] \textbf{Compute Option-wise Scores.} For each option $o$, compute:
    \[
    S(o) = 1 - \hat{p}(o)
    \]
    \textbf{Scores:} S(A) = 0.95, S(B) = 0.10, S(C) = 0.95, S(D) = 1.00

    \item[\textbf{2C.}] \textbf{Construct Prediction Set.} Include all options $o$ such that $S(o) \leq \hat{q}$.
    
    \textit{Example:} With $\hat{q} = 0.65$:
    \[
    \text{Prediction set} = \{ \text{B} \}
    \]
    Options A, C, and D are excluded as their non-conformity scores exceed the threshold.
\end{enumerate}

\subsection{Full Results}

In Tables \ref{tab:cp-llama}, \ref{tab:cp-deepseek}, \ref{tab:cp-openbio}, we report the full conformal prediction results on the test set for the three representative 70B models, broken down by medical specialty and question type. Each table reports four metrics:
\textbf{Empirical Coverage}: The proportion of test instances where the correct answer is included in the prediction set, averaged over questions in the group. This is expected to be close to the target coverage (e.g., 90\%); \textbf{Average Set Size}: The average number of answer options included in the prediction set. Lower values indicate more informative predictions; \textbf{Majority Voting Accuracy}: The accuracy of the most frequently selected answer across $M$ sampled generations (i.e., argmax over vote counts). This serves as a reference for model performance under deterministic decoding; \textbf{Hit Rate}: The proportion of test instances where at least one of the $M$ sampled generations matches the correct answer. This serves as a sort of upper-bound for empirical coverage.

These metrics are reported separately for each medical specialty and question type to highlight performance heterogeneity across clinical subdomains and reasoning tasks.

\begin{table*}[ht]
\centering
\footnotesize
\begin{tabular}{lcccc}
\toprule
\textbf{Group} & \textbf{Empirical Coverage} & \textbf{Avg Set Size} & \textbf{Maj. Voting Accuracy} & \textbf{Hit Rate} \\
\midrule
\textit{Overall}               & 85.50\% & 1.08 & 82.72\% & 95.39\% \\
Cardiology                     & 83.40\% & 1.06 & 80.69\% & 95.17\% \\
Gastroenterology              & 83.40\% & 1.11 & 82.76\% & 91.72\% \\
Infectious Diseases           & 83.30\% & 1.08 & 78.47\% & 95.83\% \\
Neurology                     & 85.40\% & 1.10 & 80.56\% & 95.14\% \\
Obstetrics/Gynecology        & 89.70\% & 1.10 & 86.90\% & 95.17\% \\
Pediatrics                    & 87.60\% & 1.04 & 86.90\% & 99.31\% \\
\midrule
Definition                    & 80.00\% & 1.00 & 80.00\% & 100.00\% \\
Diagnosis                     & 85.16\% & 1.09 & 82.32\% & 96.14\% \\
Diagnostic Test              & 86.79\% & 1.06 & 83.02\% & 92.45\% \\
Other                         & 82.61\% & 1.07 & 76.09\% & 97.83\% \\
Treatment                     & 86.77\% & 1.09 & 84.83\% & 93.77\% \\
\bottomrule
\end{tabular}
\caption{Conformal prediction results for \texttt{LLaMA-3.1-70B-instruct}. Target coverage: 90\% ($\hat{q} = 0.650$).}
\label{tab:cp-llama}
\end{table*}

\begin{table*}[ht]
\centering
\footnotesize
\begin{tabular}{lcccc}
\toprule
\textbf{Group} & \textbf{Empirical Coverage} & \textbf{Avg Set Size} & \textbf{Maj. Voting Accuracy} & \textbf{Hit Rate} \\
\midrule
\textit{Overall}               & 93.79\% & 1.18 & 89.65\% & 98.28\% \\
Cardiology                     & 94.48\% & 1.21 & 91.03\% & 97.93\% \\
Gastroenterology              & 91.03\% & 1.14 & 86.90\% & 96.55\% \\
Infectious Diseases           & 92.41\% & 1.16 & 88.97\% & 99.31\% \\
Neurology                     & 95.17\% & 1.21 & 91.03\% & 98.62\% \\
Obstetrics/Gynecology        & 93.10\% & 1.16 & 88.28\% & 99.31\% \\
Pediatrics                    & 96.55\% & 1.21 & 91.72\% & 97.93\% \\
\midrule
Definition                    & 93.33\% & 1.13 & 86.67\% & 100.00\% \\
Diagnosis                     & 95.34\% & 1.19 & 90.87\% & 98.38\% \\
Diagnostic Test              & 90.57\% & 1.09 & 86.79\% & 98.11\% \\
Other                         & 91.30\% & 1.24 & 89.13\% & 100.00\% \\
Treatment                     & 92.25\% & 1.17 & 88.37\% & 97.67\% \\
\bottomrule
\end{tabular}
\caption{Conformal prediction results for \texttt{DeepSeek-R1-70B}. Target coverage: 90\% ($\hat{q} = 0.750$).}
\label{tab:cp-deepseek}
\end{table*}

\begin{table*}[ht]
\centering
\footnotesize
\begin{tabular}{lcccc}
\toprule
\textbf{Group} & \textbf{Empirical Coverage} & \textbf{Avg Set Size} & \textbf{Maj. Voting Accuracy} & \textbf{Hit Rate} \\
\midrule
\textit{Overall}               & 90.12\% & 1.40 & 79.89\% & 96.21\% \\
Cardiology                     & 92.41\% & 1.43 & 82.07\% & 97.24\% \\
Gastroenterology              & 88.97\% & 1.38 & 77.24\% & 96.55\% \\
Infectious Diseases           & 86.21\% & 1.40 & 75.17\% & 92.41\% \\
Neurology                     & 91.03\% & 1.43 & 79.31\% & 95.86\% \\
Obstetrics/Gynecology        & 89.66\% & 1.32 & 83.45\% & 97.24\% \\
Pediatrics                    & 92.41\% & 1.42 & 82.07\% & 97.93\% \\
\midrule
Definition                    & 86.67\% & 1.27 & 80.00\% & 100.00\% \\
Diagnosis                     & 89.66\% & 1.44 & 77.28\% & 96.55\% \\
Diagnostic Test              & 94.34\% & 1.34 & 84.91\% & 98.11\% \\
Other                         & 86.96\% & 1.37 & 82.61\% & 93.48\% \\
Treatment                     & 91.09\% & 1.34 & 83.33\% & 95.74\% \\
\bottomrule
\end{tabular}
\caption{Conformal prediction results for \texttt{Llama3-OpenBioLLM-70B}. Target coverage: 90\% ($\hat{q} = 0.800$).}
\label{tab:cp-openbio}
\end{table*}

\end{document}

%% file: parts/1-new-intro-iacer.tex
\section{Introduction}
\label{sec:new_intro}

\begin{figure*}[!t]
	\centering
	\includegraphics[width=1\textwidth]{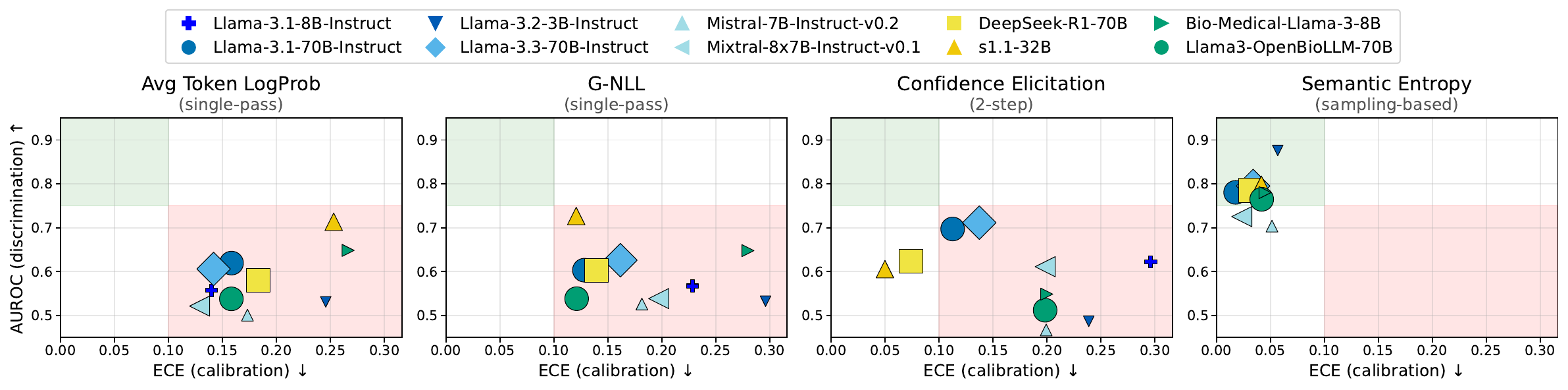}
        \caption{Uncertainty estimation for clinical QA across LLMs featuring AUROC (discrimination) vs. ECE (calibration). Marker size indicates model size; color and shape indicate model type (general, reasoning, biomedical). The green-shaded area (top left) highlights the desirable region: low calibration error and high discrimination.
        Most single-pass methods fail on both; even G-NLL~\cite{aichberger2024rethinking} underperforms despite good performance reported on non-clinical QA. Semantic Entropy achieves strong performance, though it requires multiple generations.
        }
	\label{fig:unc_plots}
\end{figure*}

In recent years, Large Language Model (LLM) performance has consistently increased across medical and clinical tasks~\cite{chen2023meditron70b,Wu2024-qf,acikgoz2024hippocrates,sellergren2025medgemmatechnicalreport}.
This includes classification tasks with finite output spaces, such as multiple-choice question answering \citep[MCQA;][]{jin2019pubmedqa,jin2021disease,pal2022medmcqa}, as well as natural language generation (NLG) tasks with open-ended output spaces~\citep{baan2023uncertainty}, such as open-ended question answering~\cite[Open-QA;][]{thirunavukarasu2023large,singhal2023large} and clinical text summarization \cite{schoonbeek_completeness_2024,van2024adapted}.
However, LLMs are known to be overconfident \citep{groot-valdenegro-toro-2024-overconfidence}, raising concerns about their reliability in high-stakes clinical settings. In such settings, reliability depends not only on \textit{discrimination} (how well confidence scores separate correct vs.~incorrect answers) but also on \textit{calibration} (how closely predicted probabilities match actual accuracy); both are critical for safe deployment \citep{atf2025challenge}.
By quantifying uncertainty, LLMs can spot unreliable predictions and prompt expert oversight if needed \citep{zhou2025uncertainty}.

Let $x \in \mathcal{X}$ be a question and $y \in \mathcal{Y}$ its answer.
Typically, in a classification setting such as MCQA, $\mathcal{Y}$ is a finite set
and quantifying the entropy of
an LLM's predictive uncertainty $p_\theta(y | x)$ for this problem can be done directly by enumerating all the possible answers.
However, uncertainty quantification (UQ) in NLG is a significant challenge~\cite{giulianelli-etal-2023-comes,geng-etal-2024-survey}.
In Open-QA, the output space $\mathcal{Y}$ is the set of all possible strings, which grows exponentially with the length of the answer. Directly calculating the predictive uncertainty $p_\theta(y | x)$ for every possible output becomes unfeasible.
In this work, we benchmark how representative UQ approaches for NLG---token-based confidence \citep{huang2025look,aichberger2024rethinking}, sample consistency \citep{kuhnsemantic}, confidence elicitation \citep{xiongcan}, and conformal prediction \citep{campos2024conformal}---fare specifically in clinical QA.
We benchmark uncertainty across \textit{10 LLMs} (general-purpose, biomedical, and reasoning-focused) and \textit{2 clinical QA datasets}, including \textit{11 different clinical specialties} (among which 5 can be considered less frequent; see Section \ref{sec:datasets} for details).
We also introduce a novel annotation scheme that maps QA pairs onto \textit{6 different question types} (e.g., diagnosis, treatment, definition questions, etc).

We also present a new case study on behavioral signals from LLM reasoning traces, where we offer a lightweight, interpretable, and single-pass alternative to sampling-based methods. We find that reasoning LLMs often yield superior accuracy and calibration, but this advantage varies significantly across specialties and question types.

We offer a cautionary yet constructive perspective on uncertainty in clinical NLP, pointing toward more context-aware evaluation. Our results show that aggregate metrics obscure important reliability differences and highlight the importance of uncertainty evaluations that account for clinical specialty and question type.
We see this work as a first step towards supporting the development of LLMs that are both accurate and trustworthy in clinical use. Our code is available at \url{https://github.com/albertotestoni/mind-the-gap-clinical-qa}.

%% file: parts/2-related.tex
\section{Related Work}
\label{sec:related}

Starting from the seminal work of \citet{guo2017calibration}, uncertainty quantification has been studied in NLP tasks such as sentiment analysis, machine translation, and named entity recognition \citep{xiao2019quantifying, zerva2022disentangling, hashimoto2024data, shorinwa2024survey}. While uncertainty estimation has a long tradition in classification tasks, the rise of LLMs has renewed interest in extending these methods to language generation \citep{geng-etal-2024-survey, ulmer-etal-2024-calibrating, liu2025uncertainty, huang2025look}. Ensuring well-calibrated predictions is essential in clinical NLP, where overconfidence or miscalibrated uncertainty estimates can translate into patient risk or diagnostic error \citep{ghoshal2021estimating, kompa2021second, dawood2023uncertainty}. 

Few approaches have systematically evaluated these aspects in the clinical domain. \citet{gao2025uncertainty} argue that LLM uncertainty signals (e.g., token-level entropy) are poorly suited for diagnosis tasks, but their analysis remains limited to a small set of outdated models and focuses solely on classification tasks. Similarly, studies on open-ended generation \citep{wu2024uncertainty, savage2025large} evaluate only a narrow range of methods and models, limiting the generalizability of their findings. Moreover, existing evaluations typically report aggregated metrics, offering limited insight into finer-grained dimensions such as medical specialties or question types. Another still underexplored aspect is the comparison across different LLM types. Recent studies have evaluated general-purpose and biomedical LLMs on clinical tasks, reporting mixed evidence for the latter's advantage \citep{dorfner2024biomedical}. However, these comparisons largely overlook uncertainty calibration. 

A recent related study to ours is \citet{savage2025large}, which investigates uncertainty estimation methods for clinical question answering. The authors found that sample consistency (assessing agreement across multiple answers generated via stochastic decoding) outperformed other approaches in terms of both discrimination and calibration, though the latter remained suboptimal. While their findings provide valuable initial insights into UQ for medical QA, their evaluation is limited in several key ways. First, it focuses on a small set of outdated models, excluding recent advances such as reasoning-tuned and domain-specific LLMs (which we highlight in Figure~\ref{fig:unc_plots}, along with the calibration–discrimination trade-off). Second, their analysis is restricted to diagnosis and treatment questions, ignoring other clinically relevant question types. Third, they do not examine how performance varies across medical specialties. A complementary line of work is \citet{ye2024benchmarking}, which applies conformal prediction to evaluate uncertainty in LLMs across five (non-clinical) NLP tasks, such as question answering and text summarization. Notably, they find that (i) higher accuracy does not necessarily correlate with higher certainty, and (ii) larger LLMs can exhibit greater uncertainty than smaller ones. In our work, we investigate whether these patterns extend to clinical question answering.

%% file: parts/3-datasets.tex
\section{Datasets and Question Annotation}
\label{sec:datasets}

We focus on two dimensions of clinical question answering that have been largely overlooked in prior work: medical specialty and question type, which induce systematic variation in task difficulty and uncertainty profiles that is obscured by aggregate evaluation.

\paragraph{By Medical Specialty}
We leverage two English multiple-choice question answering datasets. \footnote{In our setup, models generate open-ended responses, and correctness is evaluated against multiple-choice labels (§\ref{subsec:eval})}
\textbf{S-MedQA} \citep{yan2025doesneuromeancardio} combines questions from MedQA \citep{jin2021disease} and MedMCQA \citep{pal2022medmcqa}. The dataset includes a validated annotation pipeline that maps each question to one of 55 medical specialties recognized in the European Union. We consider the six most frequent specialties: \textit{Gastroenterology, Cardiology, Obstetrics and Gynecology, Neurology, Infectious Diseases, and Pediatrics}, which we refer to in the following as `common specialities'. To ensure balanced evaluation across specialties while keeping the total number of questions manageable for computational and analytical purposes, we fix the sample size to 145 questions per specialty, resulting in a total of 870 questions. Each question in S-MedQA includes four answer options and one correct answer. 
\textbf{MedExQA} \citep{kim2024medexqa} focuses on underrepresented specialities: \textit{Biomedical Engineering }(143 questions in the MedExQA test set), \textit{Clinical Laboratory Science} (372 questions), \textit{Clinical Psychology} (106 questions), \textit{Occupational Therapy} (189 questions), and \textit{Speech Language Pathology} (130 questions), summing up to 940 questions. Questions in MedExQA were manually gathered from a variety of freely accessible online sources, such as mock tests and practice exams designed for specific specialties. Each question in MedExQA includes four answer options and one correct answer.

\paragraph{By Question Type} After manually inspecting a random sample of 50 questions from each dataset, we identify six clinically meaningful question types: \textit{Diagnosis, Treatment, Diagnostic Test, Definition, Procedure/Operation}, and \textit{Other}. We annotate all 1,810 questions using \texttt{Llama-3.1-70B-Instruct} \citep{meta2024llama3.1}, and validate its outputs against manual annotations from one of the authors on a sample of 100 questions. The agreement is high, with 94\% accuracy; the few disagreements involve borderline cases of the broad \textit{Other} category. Full details on the prompt, question type definitions, and distribution are provided in Appendix~\ref{appendix:qtype}.

%% file: parts/3-methods.tex
\section{Methodology}
\label{sec:methods}

\subsection{Extracting Uncertainty Estimates}
\label{sec:extract_unc_estimates}

Uncertainty in LLMs can be quantified by exploiting the probability distributions induced by the model over its outputs. In the context of question answering, these distributions can be summarized as scalar scores or prediction sets that reflect how confident a model is in its chosen answer. We consider multiple strategies for extracting such uncertainty estimates, ranging from single-pass methods based on token probabilities to sampling-based approaches and set-valued predictions. 

\paragraph{Score-based methods:} Following the taxonomy of \citet{geng-etal-2024-survey}, we consider \textit{logit-based} approaches, such as average token-level probability under greedy decoding \citep{huang2025look}. Consistent with \citet{savage2025large}, we found negligible differences when using minimum or maximum token probability. We also include G-NLL \citep{aichberger2024rethinking}, which computes the negative log-likelihood of the greedily decoded output and serves as an efficient uncertainty proxy.\footnote{In G-NLL \citep{aichberger2024rethinking}, calibration was not included in the authors' original analysis.} For \textit{linguistic confidence} methods, we adopt a two-step confidence elicitation approach \citep{rivera-etal-2024-combining, xiongcan}, where the model first produces an answer and is then prompted to reprocess the full question–answer pair and assign a scalar certainty score. We use the same elicitation prompt as \citet{savage2025large}. 

\paragraph{Consistency-based methods:} We evaluate Semantic Entropy, inspired by \citet{kuhnsemantic, farquhar2024detecting} and related to Sample Consistency \citep{savage2025large}. While Sample Consistency measures agreement across sampled outputs, Semantic Entropy quantifies the entropy of the predicted answer distribution. Given the multiple-choice answers available in the datasets we used, we group generations by selected option and compute entropy over the resulting distribution. Higher entropy reflects greater uncertainty. %
We follow \citet{manakul-etal-2023-selfcheckgpt} and generate 10 responses per question.

\paragraph{Set-based methods:} As for a representative set-based method, we evaluate a conformal prediction approach inspired by \citet{ke2025correctness}. The method constructs prediction sets by sampling multiple responses per input and estimating empirical coverage over answer options; in our setup, we target 90\% marginal coverage of the correct answer. %
Appendix \ref{appendix:cp} provides additional details and an illustrative example.

\paragraph{Linguistic behavioral features:} We also introduce a novel case study based on \textit{linguistic behavioral features} extracted from reasoning traces, an understudied category in current taxonomies of uncertainty estimation methods for LLMs. We provide additional methodological details and analysis of this approach in Section \ref{subsec:behavioural}.

\paragraph{} Finally, following the classification in \citet{liu2025uncertainty}, we can further distinguish between \textit{single-pass} methods (logit-based scores, confidence elicitation, and linguistic behavioral features) and \textit{sampling-based} methods, which require multiple generations (consistency-based approaches). The linguistic confidence method specifically involves two passes per input.

\subsection{Models}
We evaluate 10 open-source instruction-tuned LLMs to comprehensively estimate uncertainty across model architecture, scale, and domain specialization. 
We include six \textbf{general-purpose} models of varying sizes and architectures: \texttt{LLaMA-3.1-8B}, \texttt{LLaMA-3.1-70B}, \texttt{LLaMA-3.2-3B}, \texttt{LLaMA-3.3-70B} \citep{meta2024llama3}, \texttt{Mistral-7B-v0.2} and \texttt{Mixtral-8x7B-v0.1} \citep{mistral2023mistral}; two \textbf{reasoning-oriented} models: \texttt{DeepSeek-R1-70B} \citep{guo2025deepseek} and \texttt{s1.1-32B} \citep{muennighoff2025s1}; and two \textbf{domain-specific biomedical} models: \texttt{Bio-Medical-LLaMA-3-8B} \citep{ContactDoctor_Bio-Medical-Llama-3-8B} and \texttt{LLaMA3-OpenBioLLM-70B} \citep{OpenBioLLMs}, both fine-tuned on biomedical and clinical corpora.
In Appendix \ref{appendix:models}, \ref{appendix:prompts}, and \ref{appendix:decoding} we provide details about the models, prompts, and decoding hyperparameters, respectively. Appendix \ref{appendix:gpt} presents a focused analysis on three \textbf{OpenAI ChatGPT models} (\texttt{GPT-4.1}, \texttt{GPT-4.1-mini}, and \texttt{GPT-4o-mini}).

\subsection{Evaluation}
\label{subsec:eval}

\paragraph{Answer Extraction and Accuracy} 
Models are prompted to output a long-form answer that includes the letter of the selected option in square brackets (e.g., \texttt{[C]}), which we extract via regular expressions. Only when the format deviates (on average in ~5–6\% of responses, e.g., full answer text within brackets instead of the letter alone), we use \texttt{LLaMA-3.1-8B-Instruct} to map the response to one of the four options (see Appendix \ref{appendix:prompts} for additional details). We validate its performance on this task against human annotations over 100 samples, observing 92\% accuracy. \textit{QA accuracy is computed by comparing the extracted answer to the ground-truth label}, whereas \textit{uncertainty estimates are computed over the full (long-form) answer generated by the LLM}. To control for position bias, answer options are shuffled four times per question with greedy decoding. Semantic Entropy and Conformal Prediction require sampling (via nucleus sampling) 10 and 20 responses per question, respectively, with the options shuffled at each generation to mitigate position bias. 

\paragraph{Discrimination and Calibration} 
To evaluate discrimination, i.e., the ability of uncertainty estimates to distinguish correct from incorrect predictions, we compute the Area Under the Receiver Operating Characteristic Curve (AUROC). AUROC measures how well uncertainty scores separate correct from incorrect predictions; a value of 1.0 indicates perfect discrimination, while 0.5 corresponds to random guessing. We assess calibration, i.e., the alignment between predicted confidence and empirical accuracy, using Expected Calibration Error (ECE) and the Brier Score \citep{glenn1950verification}. ECE measures the discrepancy between predicted confidence and actual accuracy. Following \citet{rivera-etal-2024-combining}, we use 10 bins to compute ECE. The Brier Score captures the squared error between predicted confidence and correctness, providing a proper scoring rule that reflects both calibration and sharpness. For both scores, a value of 0 indicates perfect calibration. Given that ECE and Brier Score rank models similarly, we report their average for some of the analyses below.

\begin{table*}[t]
	\centering
	\scriptsize
	\renewcommand{\arraystretch}{1.2}
	\setlength{\tabcolsep}{2pt}
	\begin{tabular}{rccccccccccc}
		\hline
		Model & \rotatebox{45}{Gastro.} & \rotatebox{45}{Cardiology} & \rotatebox{45}{Obs. and gyn.} & \rotatebox{45}{Neurology} & \rotatebox{45}{Infectious dis.} & \rotatebox{45}{Pediatrics} & \rotatebox{45}{Biomed. Eng.} & \rotatebox{45}{Clinical Lab.} & \rotatebox{45}{Clinical Psy.} & \rotatebox{45}{Occ. Therapy} & \rotatebox{45}{\shortstack{Speech-Lang.~~ \\Path.}} \\ \hline
		\hline
		Llama-3.1-8B & \cellcolor[RGB]{234,241,237} 0.623 & \cellcolor[RGB]{227,238,231} 0.633 & \cellcolor[RGB]{214,233,222} 0.651 & \cellcolor[RGB]{237,242,239} 0.617 & \cellcolor[RGB]{246,246,246} 0.601 & \cellcolor[RGB]{237,242,239} 0.616 & \cellcolor[RGB]{171,214,190} 0.714 & \cellcolor[RGB]{152,206,175} 0.742 & \cellcolor[RGB]{124,194,154} 0.785 & \cellcolor[RGB]{124,194,154} 0.784 & \cellcolor[RGB]{244,168,134} 0.286 \\
		Llama-3.1-70B & \cellcolor[RGB]{135,198,162} \textbf{0.769} & \cellcolor[RGB]{126,195,155} 0.781 & \cellcolor[RGB]{95,181,132} \textbf{0.828} & \cellcolor[RGB]{131,197,159} 0.774 & \cellcolor[RGB]{147,204,171} \textbf{0.750} & \cellcolor[RGB]{81,175,121} \textbf{0.848} & \cellcolor[RGB]{67,169,111} \textbf{0.868} & \cellcolor[RGB]{55,164,102} 0.887 & \cellcolor[RGB]{29,153,82} \textbf{0.925} & \cellcolor[RGB]{67,169,111} \textbf{0.870} & \cellcolor[RGB]{244,181,153} \textbf{0.342} \\
		Llama-3.2-3B & \cellcolor[RGB]{246,230,223} 0.538 & \cellcolor[RGB]{245,217,204} 0.484 & \cellcolor[RGB]{246,224,214} 0.512 & \cellcolor[RGB]{246,231,225} 0.543 & \cellcolor[RGB]{246,229,222} 0.534 & \cellcolor[RGB]{246,224,214} 0.512 & \cellcolor[RGB]{246,239,236} 0.572 & \cellcolor[RGB]{208,230,217} 0.661 & \cellcolor[RGB]{204,228,214} 0.665 & \cellcolor[RGB]{173,215,191} 0.713 & \cellcolor[RGB]{244,167,133} 0.285 \\
		Llama-3.3-70B & \cellcolor[RGB]{135,198,162} \textbf{0.769} & \cellcolor[RGB]{105,186,140} \textbf{0.812} & \cellcolor[RGB]{107,187,141} 0.809 & \cellcolor[RGB]{168,213,187} 0.719 & \cellcolor[RGB]{149,204,172} 0.748 & \cellcolor[RGB]{135,198,162} 0.769 & \cellcolor[RGB]{91,180,129} 0.834 & \cellcolor[RGB]{58,166,104} 0.881 & \cellcolor[RGB]{79,175,120} 0.851 & \cellcolor[RGB]{81,175,121} 0.848 & \cellcolor[RGB]{244,180,152} \textbf{0.335} \\
		Mistral-7B & \cellcolor[RGB]{245,211,196} 0.463 & \cellcolor[RGB]{246,228,220} 0.529 & \cellcolor[RGB]{246,222,212} 0.505 & \cellcolor[RGB]{245,209,193} 0.452 & \cellcolor[RGB]{245,196,174} 0.401 & \cellcolor[RGB]{245,206,188} 0.440 & \cellcolor[RGB]{123,193,153} 0.787 & \cellcolor[RGB]{181,219,197} 0.700 & \cellcolor[RGB]{138,200,165} 0.764 & \cellcolor[RGB]{140,201,166} 0.762 & \cellcolor[RGB]{244,168,134} 0.287 \\
		Mixtral-8x7B & \cellcolor[RGB]{246,244,242} 0.590 & \cellcolor[RGB]{244,245,245} 0.607 & \cellcolor[RGB]{221,236,228} 0.641 & \cellcolor[RGB]{246,239,236} 0.573 & \cellcolor[RGB]{246,224,214} 0.512 & \cellcolor[RGB]{246,242,240} 0.584 & \cellcolor[RGB]{107,187,141} 0.811 & \cellcolor[RGB]{117,191,149} 0.794 & \cellcolor[RGB]{90,179,128} 0.835 & \cellcolor[RGB]{119,192,150} 0.792 & \cellcolor[RGB]{244,167,133} 0.285 \\
		DeepSeek-R1-70B & \cellcolor[RGB]{95,181,132} \textbf{0.828} & \cellcolor[RGB]{91,180,129} \textbf{0.833} & \cellcolor[RGB]{86,178,125} \textbf{0.841} & \cellcolor[RGB]{93,181,131} \textbf{0.831} & \cellcolor[RGB]{72,172,115} \textbf{0.862} & \cellcolor[RGB]{55,164,102} \textbf{0.886} & \cellcolor[RGB]{62,167,107} \textbf{0.877} & \cellcolor[RGB]{43,159,93} \textbf{0.904} & \cellcolor[RGB]{26,152,80} \textbf{0.932} & \cellcolor[RGB]{76,173,117} 0.856 & \cellcolor[RGB]{244,174,142} 0.310 \\
		s1.1-32B & \cellcolor[RGB]{83,176,123} \textbf{0.847} & \cellcolor[RGB]{83,176,123} \textbf{0.845} & \cellcolor[RGB]{79,175,120} \textbf{0.850} & \cellcolor[RGB]{100,184,136} \textbf{0.819} & \cellcolor[RGB]{69,170,112} \textbf{0.867} & \cellcolor[RGB]{69,170,112} \textbf{0.867} & \cellcolor[RGB]{84,177,124} 0.844 & \cellcolor[RGB]{48,161,97} \textbf{0.896} & \cellcolor[RGB]{29,153,82} \textbf{0.925} & \cellcolor[RGB]{72,172,115} \textbf{0.861} & \cellcolor[RGB]{244,175,145} 0.318 \\
		Llama3-OpenBioLLM-70B & \cellcolor[RGB]{168,213,187} 0.719 & \cellcolor[RGB]{142,201,167} 0.757 & \cellcolor[RGB]{142,201,167} 0.759 & \cellcolor[RGB]{130,196,158} \textbf{0.777} & \cellcolor[RGB]{166,212,186} 0.723 & \cellcolor[RGB]{128,195,157} 0.779 & \cellcolor[RGB]{60,166,106} \textbf{0.880} & \cellcolor[RGB]{48,161,97} \textbf{0.897} & \cellcolor[RGB]{48,161,97} 0.897 & \cellcolor[RGB]{71,171,114} \textbf{0.865} & \cellcolor[RGB]{244,182,154} \textbf{0.345} \\
		Bio-Medical-Llama-3-8B & \cellcolor[RGB]{213,232,221} 0.653 & \cellcolor[RGB]{216,233,224} 0.647 & \cellcolor[RGB]{178,217,195} 0.705 & \cellcolor[RGB]{216,233,224} 0.648 & \cellcolor[RGB]{209,230,218} 0.659 & \cellcolor[RGB]{220,235,226} 0.643 & \cellcolor[RGB]{152,206,175} 0.743 & \cellcolor[RGB]{166,212,186} 0.722 & \cellcolor[RGB]{102,184,137} 0.818 & \cellcolor[RGB]{156,207,178} 0.737 & \cellcolor[RGB]{244,165,130} 0.273 \\
		\hline
	\end{tabular}
	\caption{Multiple-choice QA accuracy by medical specialty, revealing substantial heterogeneity in model performance across domains. For each specialty, the three best-performing models are highlighted in bold.}
	\label{tab:acc_per_specialty_rwgreen}
\end{table*}

\paragraph{Conformal Prediction}
For Conformal Prediction, we evaluate uncertainty in terms of both empirical coverage (the proportion of test instances for which the correct answer is included in the prediction set) and average set size (how many options are included on average in the prediction set). We target a nominal coverage level of 90\%, and report the realized coverage. Intuitively, a lower average set size indicates more confident (i.e., less uncertain) predictions, while higher empirical coverage reflects greater reliability. However, the two must be balanced: low set size with low coverage suggests overconfidence, while high coverage with large sets indicates conservativeness.

%% file: parts/4-results.tex
\section{Score-based UQ Methods Results}
\label{sec:results}

\begin{figure*}[t]
	\centering
	\includegraphics[width=1\textwidth]{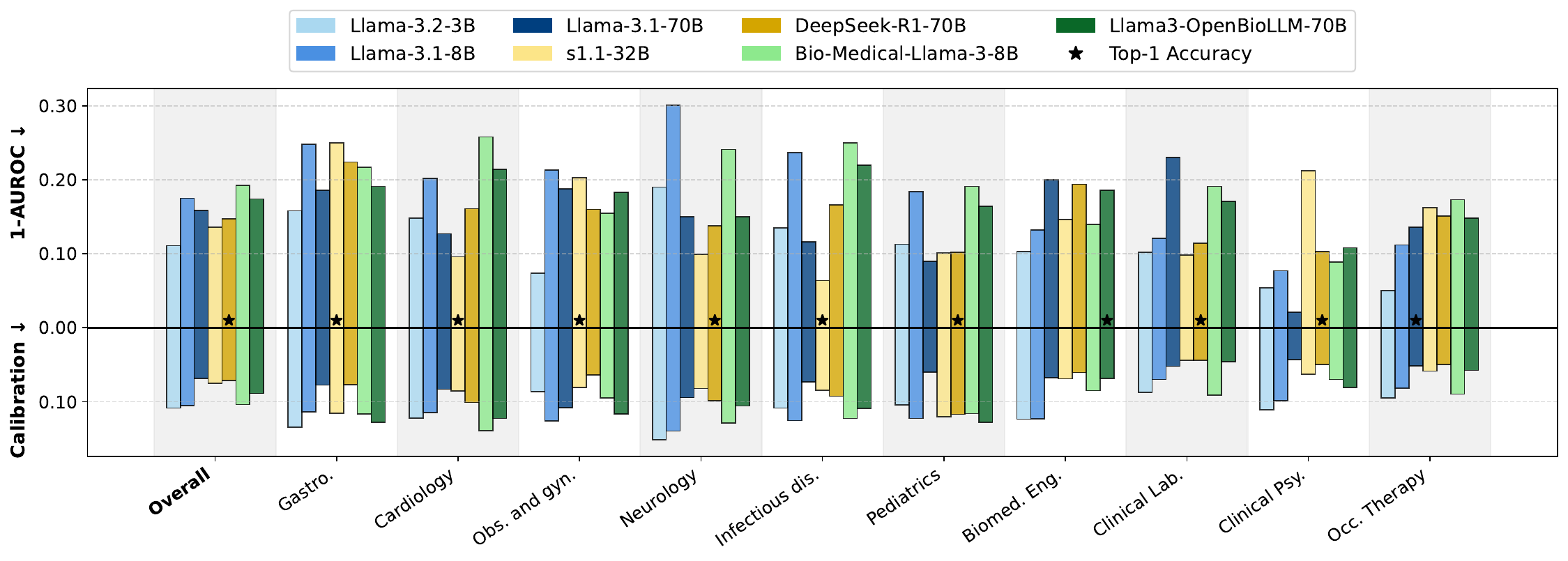}
	\caption{Discrimination (1$-$AUROC) and calibration (average ECE and Brier score) of semantic entropy–based uncertainty estimates across specialties. $\star$ indicates the most accurate model per specialty.}
	\label{fig:plot_discrimination_calibration}
\end{figure*}

We begin by evaluating score-based uncertainty methods by analyzing how their scores relate to prediction correctness, focusing on calibration and discrimination across all models and questions.

\subsection{From Logits to Semantic Entropy}
\label{subsec:scalar_uq}
We compare logit-based, confidence elicitation, and consistency-based methods to assess (a) which yields the most calibrated and discriminative estimates, and (b) whether LLM type or size systematically affect them. Figure~\ref{fig:unc_plots} shows AUROC vs.~ECE for all LLMs and UQ methods. 
In line with the findings in \citet{savage2025large}, \textit{Semantic Entropy} shows the strongest overall performance while requiring multiple generations per input. Reasoning-oriented and biomedical LLMs do not consistently outperform general-purpose models in terms of uncertainty estimates. G-NLL does not perform well in either dimension, although it performs better on the recent \texttt{s1.1-32B} reasoning model. In general, larger models tend to yield more reliable uncertainty estimates (particularly for Confidence Elicitation), though this trend is less pronounced for Semantic Entropy, which performs robustly across nearly all models, including the smallest (\texttt{LLaMA-3.2-3B}). By contrast, \texttt{Mistral-7B} and \texttt{Mixtral-8x7B} underperform relative to the rest.

Based on these findings, we adopt Semantic Entropy as our primary uncertainty estimation method for the remainder of this section. We revisit single-generation alternatives in Section~\ref{subsec:behavioural}. While Semantic Entropy provides well-calibrated uncertainty estimates across a broad range of models, two important questions remain:
\textit{Does this performance hold uniformly across medical specialties and question types?}
And \textit{how are these uncertainty estimates aligned with actual QA accuracy?}

\subsubsection{Impact of Medical Specialty}
\paragraph{Model Accuracy:} Table~\ref{tab:acc_per_specialty_rwgreen} reports model accuracy across medical specialties, with the top three models per specialty highlighted in bold. We observe a consistent effect of model size: larger models generally achieve higher accuracy than their smaller counterparts, particularly within the \texttt{LLaMA} family. Reasoning models (\texttt{s1.1-32B} and \texttt{DeepSeek-R1-70B}), trained on reasoning traces, frequently rank among the top performers, suggesting that prior exposure to reasoning data contributes to robust performance across domains. Biomedical models tend to outperform general-purpose models at smaller scales (e.g., 8B), but this advantage does not hold at larger scales, where models such as \texttt{LLaMA-3.1-70B} surpass their domain-specific counterparts (e.g., \texttt{Llama3-OpenBioLLM-70B}). Notably, all models perform only marginally above chance on  \textit{Speech-Language Pathology}, echoing findings from the original dataset authors \citep{kim2024medexqa}. 
Overall, these observations raise the question of whether more accurate models are also more calibrated or reliable in their predictions, and how this aligns with findings by \citet{ye2024benchmarking}, who report that larger-scale LLMs may exhibit greater uncertainty.

\paragraph{Discrimination vs Calibration:} Figure~\ref{fig:plot_discrimination_calibration} shows discrimination (1–AUROC) and calibration (avg. ECE and Brier score) for a representative set of models. While larger models generally achieve higher accuracy (no model $<$ 32B ranked top-3 in any specialty), this does not guarantee superior uncertainty estimates. %
For instance, \texttt{Llama-3.2-3B} achieves the highest AUROC, significantly outperforming the second-best model, \texttt{s1.1-32B} ($p=0.03$, bootstrap test). However, it simultaneously ranks among the worst in calibration, highlighting a clear decoupling between discrimination and calibration.
Interesting trade-offs emerge between accuracy and calibration, e.g., in \textit{Pediatrics}, \texttt{DeepSeek-R1-70B} achieves the highest accuracy (\textasciitilde89\%) but has worse calibration than \texttt{Llama-3.1-70B} ($p=0.05$, Brier score), whose accuracy is slightly lower (\textasciitilde85\%) but much better calibrated. This suggests that gains in accuracy can come at the cost of overconfidence.

\paragraph{Model Specialization:}  Smaller models occasionally demonstrate strong discrimination despite low accuracy. In \textit{Obstetrics and Gynecology}, \texttt{Llama-3.2-3B} answers just \textasciitilde51\% correctly but achieves the best AUROC, suggesting it is reliably uncertain when unsure. Conversely, some larger models show worse AUROC despite higher accuracy, indicating overconfidence on their few errors. \texttt{DeepSeek-R1-70B} consistently exhibits strong calibration, achieving the lowest average ECE and Brier scores in about half the specialties. Meanwhile, domain-specialized models such as \texttt{Llama3-OpenBioLLM-70B} excel in some focus areas (e.g., \textit{Biomedical Engineering}) but underperform on general domains like \textit{Cardiology} or \textit{Infectious Disease}, highlighting the limited generalizability of domain tuning. Together, these results underscore that high performance in one evaluation dimension (accuracy, discrimination, calibration) does not imply strength in the others, reinforcing the need to evaluate all three dimensions for safe deployment.

\begin{table}[t]
\centering
\begin{adjustbox}{max width=\linewidth}
\renewcommand{\arraystretch}{1.2}
\setlength{\tabcolsep}{2pt}
\begin{tabular}{rcccccc}
\hline
Model & \rotatebox{58}{Diagnosis} & \rotatebox{58}{Treatment} & \rotatebox{58}{\shortstack{Diagnostic \\Test}} & \rotatebox{58}{Definition} & \rotatebox{58}{\shortstack{Procedure/\\Operation}} & \rotatebox{58}{Other} \\ \hline
\hline
Llama-3.1-8B & \cellcolor[RGB]{246,233,227} 0.620 & \cellcolor[RGB]{246,242,241} 0.646 & \cellcolor[RGB]{244,245,245} 0.659 & \cellcolor[RGB]{169,213,188} 0.730 & \cellcolor[RGB]{246,232,226} 0.618 & \cellcolor[RGB]{245,209,193} 0.558 \\
Llama-3.1-70B & \cellcolor[RGB]{107,187,141} \textbf{0.790} & \cellcolor[RGB]{126,195,155} 0.772 & \cellcolor[RGB]{78,174,119} 0.818 & \cellcolor[RGB]{41,158,91} \textbf{0.853} & \cellcolor[RGB]{214,233,222} \textbf{0.686} & \cellcolor[RGB]{166,212,186} 0.733 \\
Llama-3.2-3B & \cellcolor[RGB]{245,201,182} 0.538 & \cellcolor[RGB]{245,203,185} 0.543 & \cellcolor[RGB]{245,213,199} 0.570 & \cellcolor[RGB]{246,242,241} 0.645 & \cellcolor[RGB]{244,165,130} 0.443 & \cellcolor[RGB]{244,166,132} 0.448 \\
Llama-3.3-70B & \cellcolor[RGB]{131,197,159} 0.766 & \cellcolor[RGB]{136,199,163} 0.761 & \cellcolor[RGB]{91,180,129} 0.805 & \cellcolor[RGB]{60,166,106} 0.835 & \cellcolor[RGB]{246,238,234} \textbf{0.633} & \cellcolor[RGB]{213,232,221} 0.688 \\
Mistral-7B-v0.2 & \cellcolor[RGB]{244,179,151} 0.482 & \cellcolor[RGB]{244,188,163} 0.503 & \cellcolor[RGB]{246,232,226} 0.618 & \cellcolor[RGB]{180,218,196} 0.720 & \cellcolor[RGB]{245,205,187} 0.549 & \cellcolor[RGB]{245,198,177} 0.530 \\
Mixtral-8x7B-v0.1 & \cellcolor[RGB]{246,220,209} 0.588 & \cellcolor[RGB]{246,224,215} 0.598 & \cellcolor[RGB]{181,219,197} 0.719 & \cellcolor[RGB]{121,192,152} 0.776 & \cellcolor[RGB]{246,229,221} 0.611 & \cellcolor[RGB]{246,231,224} 0.616 \\
DeepSeek-R1-70B & \cellcolor[RGB]{57,165,103} \textbf{0.839} & \cellcolor[RGB]{90,179,128} \textbf{0.806} & \cellcolor[RGB]{31,154,83} \textbf{0.863} & \cellcolor[RGB]{41,158,91} \textbf{0.853} & \cellcolor[RGB]{246,237,232} 0.631 & \cellcolor[RGB]{126,195,155} \textbf{0.772} \\
s1.1-32B & \cellcolor[RGB]{46,160,95} \textbf{0.848} & \cellcolor[RGB]{102,184,137} \textbf{0.795} & \cellcolor[RGB]{62,167,107} \textbf{0.833} & \cellcolor[RGB]{50,162,98} 0.845 & \cellcolor[RGB]{246,231,225} 0.617 & \cellcolor[RGB]{78,174,119} \textbf{0.819} \\
Llama3-OpenBioLLM-70B & \cellcolor[RGB]{161,210,182} 0.739 & \cellcolor[RGB]{123,193,153} \textbf{0.775} & \cellcolor[RGB]{71,171,114} \textbf{0.825} & \cellcolor[RGB]{26,152,80} \textbf{0.869} & \cellcolor[RGB]{185,220,200} \textbf{0.715} & \cellcolor[RGB]{133,198,161} \textbf{0.765} \\
Bio-Medical-Llama-3-8B & \cellcolor[RGB]{246,246,246} 0.657 & \cellcolor[RGB]{246,240,237} 0.639 & \cellcolor[RGB]{246,235,230} 0.627 & \cellcolor[RGB]{162,210,183} 0.737 & \cellcolor[RGB]{246,221,210} 0.591 & \cellcolor[RGB]{246,224,214} 0.597 \\
\hline
\end{tabular}
\end{adjustbox}
\caption{Accuracy per question type. Top-3 models per question type are shown in bold.}
\label{tab:acc_per_qtype_rwgreen}
\end{table}

\begin{figure}[t]
  \centering
  \includegraphics[width=\linewidth]{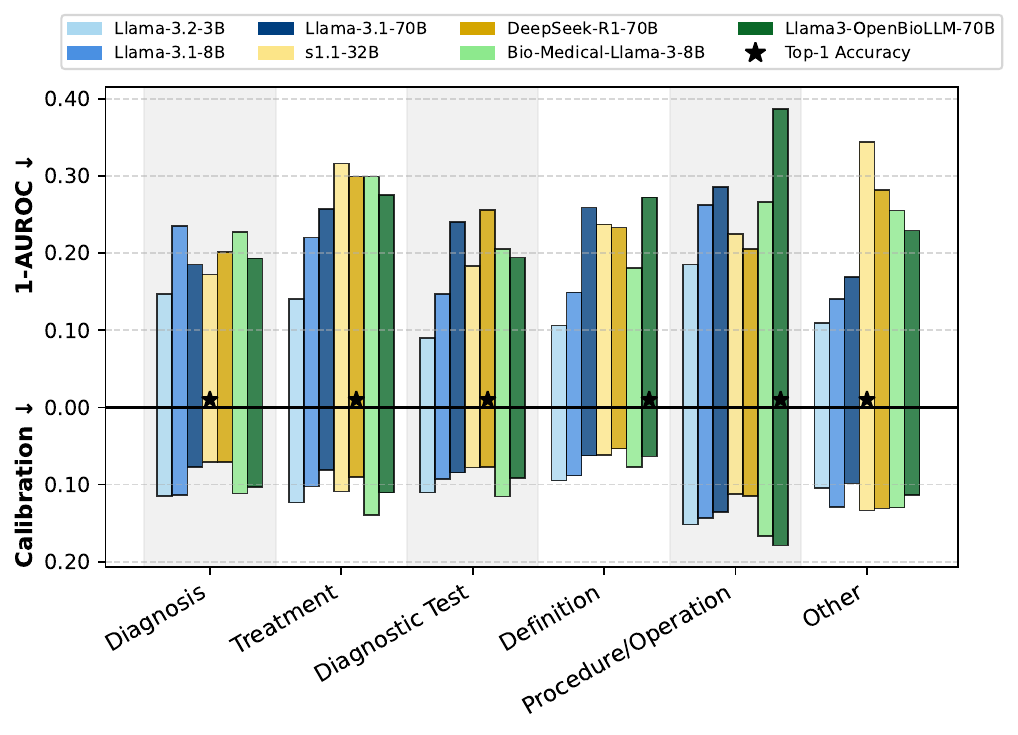}
  \caption{Calibration/discrimination per question type using Semantic Entropy estimates.}
  \label{fig:plot_dc_score_qtype}
\end{figure}

\subsubsection{Impact of Question Type}

Table~\ref{tab:acc_per_qtype_rwgreen} reports accuracy by question type, leveraging the annotation pipeline introduced in Section \ref{sec:datasets}. Several trends observed in the per-specialty analysis persist: larger \texttt{LLaMA} models consistently outperform their smaller counterparts, and reasoning-oriented models frequently rank among the top in accuracy. Our new question-type annotations also reveal \textit{Procedure/Operation} as a particularly challenging category, with all models underperforming and only \texttt{OpenBioLLM-70B} exceeding 70\% accuracy. 
By contrast, \textit{Definition} questions emerge as the easiest category, with nearly all models achieving high accuracy. This may reflect the fact that definition-style queries are closer in form and content to pretraining data, often resembling glossary entries or encyclopedic descriptions.

\begin{figure*}[t]
	\centering
	\includegraphics[width=1\linewidth]{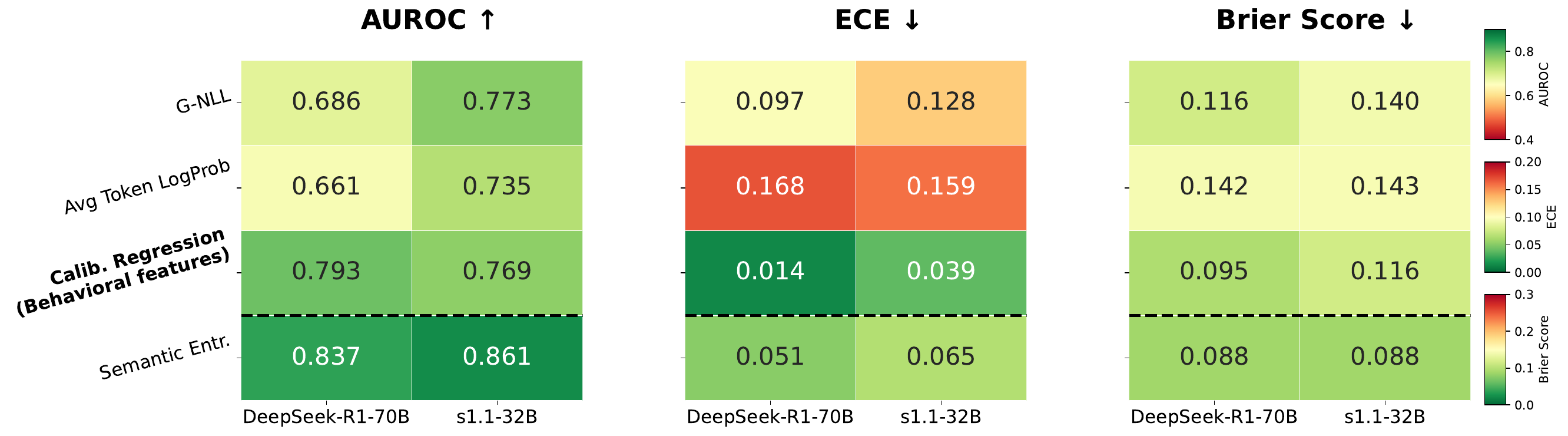}
	\caption{Our proposed method, based on behavioral features in reasoning-oriented LLMs (shown in bold), performs strongly across AUROC, ECE, and Brier Score, approaching Semantic Entropy, which instead relies on multiple generations via sampling.
	}
	\label{fig:ling_features_smedqa}
\end{figure*}

The quality of uncertainty estimates varies substantially across question types (Figure~\ref{fig:plot_dc_score_qtype}). On harder question types (in terms of accuracy), models not only make more errors but also exhibit weaker confidence separation between correct and incorrect answers. Interestingly, the smallest model (3B) achieves the best discrimination across all question types despite the lowest accuracy and poor calibration, confirming the pattern observed for medical specialties. No single model dominates across all types: \texttt{Llama3-OpenBioLLM-70B} achieves the highest accuracy on the challenging \textit{Procedure/Operation} questions, yet also exhibits the worst calibration and discrimination on the same category, suggesting that its confident predictions are often misaligned with actual correctness.

\subsection{Behavioural Signals of Uncertainty in Reasoning Models: A Case Study}
\label{subsec:behavioural}

Although performance varies across medical specialties and question types, Semantic Entropy substantially outperforms sentence-level logprob methods in both calibration and discrimination. While effective, the method requires repeated generations: this feature may hinder its deployment in real-world applications, particularly in latency-sensitive or resource-constrained settings, such as clinical decision support. In contrast, single-pass methods like G-NLL and average token probability require only one forward pass per input, albeit with reduced discrimination and calibration performance.

To address this trade-off, we explore whether behavioral signals in LLM reasoning traces can serve as lightweight proxies for uncertainty. Our motivation stems from empirical observations: models like \texttt{DeepSeek-R1-70B} and \texttt{s1.1-32B} often produce intermediate reasoning steps with linguistic cues that appear to reflect uncertainty. For example, we frequently observe self-directed questions and tokens such as ``Wait\ldots'', which may signal self-verification or reconsideration. We also hypothesize that longer responses may indicate greater difficulty in reaching a final answer. Appendix~\ref{appendix:behavioural} provides illustrative examples of these patterns. Notably, the authors of \texttt{s1.1-32B} show that ``Wait'' tokens can be leveraged at test time to improve prediction calibration \citep{muennighoff2025s1}.

\begin{table*}[t]
	\centering
	\renewcommand{\arraystretch}{1.3}
	\setlength{\tabcolsep}{3pt}
	\begin{adjustbox}{max width=\textwidth}
		\begin{tabular}{l|c|*{6}{c}|*{5}{c}}
			\toprule
			\textbf{Model} & \textbf{Overall} & \textbf{Cardio} & \textbf{Gastro} & \textbf{Inf. Dis.} & \textbf{Neuro} & \textbf{Ob/Gyn} & \textbf{Peds} & 
			\textbf{Def.} & \textbf{Diag.} & \textbf{Dx Test} & \textbf{Treat.} & \textbf{Other} \\
			\midrule
			\textbf{Llama-3.1-70B} &
			\makecell{\textcolor{red!70!black}{85.5\%}\\\textcolor{blue!18!purple}{(1.08)}} &
			\makecell{\textcolor{red!70!black}{83.4\%}\\\textcolor{blue!14!purple}{(1.06)}} &
			\makecell{\textcolor{red!70!black}{83.4\%}\\\textcolor{blue!25!purple}{(1.11)}} &
			\makecell{\textcolor{red!70!black}{83.3\%}\\\textcolor{blue!18!purple}{(1.08)}} &
			\makecell{\textcolor{red!70!black}{85.4\%}\\\textcolor{blue!23!purple}{(1.10)}} &
			\makecell{\textcolor{red!70!black}{89.7\%}\\\textcolor{blue!23!purple}{(1.10)}} &
			\makecell{\textcolor{red!70!black}{87.6\%}\\\textcolor{blue!9!purple}{(1.04)}} &
			\makecell{\textcolor{red!70!black}{80.0\%}\\\textcolor{blue!0!purple}{(1.00)}} &
			\makecell{\textcolor{red!70!black}{85.2\%}\\\textcolor{blue!20!purple}{(1.09)}} &
			\makecell{\textcolor{red!70!black}{86.8\%}\\\textcolor{blue!14!purple}{(1.06)}} &
			\makecell{\textcolor{red!70!black}{86.8\%}\\\textcolor{blue!20!purple}{(1.09)}} &
			\makecell{\textcolor{red!70!black}{82.6\%}\\\textcolor{blue!16!purple}{(1.07)}} \\
			\midrule
			\textbf{DeepSeek-R1-70B} &
			\makecell{\textcolor{green!50!black}{93.8\%}\\\textcolor{blue!41!purple}{(1.18)}} &
			\makecell{\textcolor{green!50!black}{94.5\%}\\\textcolor{blue!48!purple}{(1.21)}} &
			\makecell{\textcolor{green!50!black}{91.0\%}\\\textcolor{blue!32!purple}{(1.14)}} &
			\makecell{\textcolor{green!50!black}{92.4\%}\\\textcolor{blue!36!purple}{(1.16)}} &
			\makecell{\textcolor{green!50!black}{95.2\%}\\\textcolor{blue!48!purple}{(1.21)}} &
			\makecell{\textcolor{green!50!black}{93.1\%}\\\textcolor{blue!36!purple}{(1.16)}} &
			\makecell{\textcolor{green!50!black}{96.5\%}\\\textcolor{blue!48!purple}{(1.21)}} &
			\makecell{\textcolor{green!50!black}{93.3\%}\\\textcolor{blue!30!purple}{(1.13)}} &
			\makecell{\textcolor{green!50!black}{95.3\%}\\\textcolor{blue!43!purple}{(1.19)}} &
			\makecell{\textcolor{green!50!black}{90.6\%}\\\textcolor{blue!20!purple}{(1.09)}} &
			\makecell{\textcolor{green!50!black}{92.2\%}\\\textcolor{blue!39!purple}{(1.17)}} &
			\makecell{\textcolor{green!50!black}{91.3\%}\\\textcolor{blue!55!purple}{(1.24)}} \\
			\midrule
			\textbf{Llama3-OpenBioLLM-70B} &
			\makecell{\textcolor{green!50!black}{90.1\%}\\\textcolor{blue!91!purple}{(1.40)}} &
			\makecell{\textcolor{green!50!black}{92.4\%}\\\textcolor{blue!98!purple}{(1.43)}} &
			\makecell{\textcolor{red!70!black}{89.0\%}\\\textcolor{blue!86!purple}{(1.38)}} &
			\makecell{\textcolor{red!70!black}{86.2\%}\\\textcolor{blue!91!purple}{(1.40)}} &
			\makecell{\textcolor{green!50!black}{91.0\%}\\\textcolor{blue!98!purple}{(1.43)}} &
			\makecell{\textcolor{red!70!black}{89.7\%}\\\textcolor{blue!73!purple}{(1.32)}} &
			\makecell{\textcolor{green!50!black}{92.4\%}\\\textcolor{blue!95!purple}{(1.42)}} &
			\makecell{\textcolor{red!70!black}{86.7\%}\\\textcolor{blue!61!purple}{(1.27)}} &
			\makecell{\textcolor{red!70!black}{89.7\%}\\\textcolor{blue!100!purple}{(1.44)}} &
			\makecell{\textcolor{green!50!black}{94.3\%}\\\textcolor{blue!77!purple}{(1.34)}} &
			\makecell{\textcolor{green!50!black}{91.1\%}\\\textcolor{blue!77!purple}{(1.34)}} &
			\makecell{\textcolor{red!70!black}{87.0\%}\\\textcolor{blue!84!purple}{(1.37)}} \\
			\bottomrule
		\end{tabular}
	\end{adjustbox}
	\caption{Conformal Prediction. Empirical coverage (top) and average set size (bottom, in parentheses) for each model. Coverage is colored green ($\geq$90\%) or red (<90\%). Set size transitions from blue to purple based on value.}
    \label{tab:conformal}
\end{table*}

To operationalize these hypotheses, we extract three surface-level behavioral features from each model response: (1) the number of tokens in the reasoning trace, (2) the number of self-posed questions, and (3) the number of occurrences of the token ``Wait''. We explore different strategies for turning these features into uncertainty estimates, detailed in Appendix \ref{appendix:behavioural}. The most successful method treats the three features as input to a Calibration Regression supervised model (which fits a linear function with L2 regularization as \textit{Ridge Regression}). To train this model, we use ground-truth correctness labels (1 for correct, 0 for incorrect) for each response. Although the regression model is trained to predict correctness, we hypothesize that its outputs can serve as proxies for confidence. We define uncertainty as its complement, \( u = 1 - f(\mathbf{x}) \), enabling direct comparison with other scalar-based uncertainty estimation methods. To ensure fair comparison and avoid overfitting, we train the regression models on 40\% of the dataset and evaluate on the remaining 60\%, preserving the original distribution of medical specialties. We assess the quality of the behavioral uncertainty signals using AUROC, ECE, and Brier Score. To further disentangle model behavior, we report results separately for common and uncommon medical specialties (see Appendix~\ref{appendix:behavioural} for the latter). We include results with Semantic Entropy for comparison, noting that it relies on repeated sampling. 

The results, shown in Figure \ref{fig:ling_features_smedqa}, reveal that our lightweight behavioral feature method outperforms sentence-level methods and closely matches Semantic Entropy, while requiring only a single forward pass per input. This pattern holds across both reasoning-oriented LLMs, \texttt{DeepSeek-R1-70B} and \texttt{s1.1-32B}, and applies to both calibration and discrimination metrics. In Appendix \ref{appendix:regression_coefficients}, we provide further details on the regression model coefficients. Although restricted to two models, these results support our hypothesis that behavioral cues in reasoning traces encode relevant information about model uncertainty and highlight a promising direction for efficient, lightweight, and interpretable uncertainty estimation without requiring sampling or token-level probabilities. This has important implications for real-world applications, where uncertainty needs to be estimated reliably and efficiently. Our case study offers a starting point for future work to validate this approach across broader range of tasks and models. We view our contribution as an initial step toward establishing a more general framework for LLM behavioral uncertainty estimation leveraging structured reasoning traces.

%% file: parts/5-CF.tex
\section{Conformal Prediction Results}
\label{sec:cf}

Analyses based on score-based uncertainty estimates (Section~\ref{sec:results}) reveal notable trends: reasoning LLMs often achieve higher accuracy and better calibration, while biomedical models often appear overconfident (likely due to domain-specific tuning) and underperform outside a narrow range of specialties, leading to less reliable estimates.

We next examine conformal prediction (following the method in \citealt{ke2025correctness}) to assess whether set-based uncertainty estimates offer complementary insights. We focus on three representative 70B LLMs (one per model type) and restrict the analysis to common medical specialties with sufficient held-out data. Table~\ref{tab:conformal} reports empirical coverage and average set size across models, specialties, and question types. The three models achieve strong overall coverage, though not all of them reach the 90\% coverage target. \texttt{DeepSeek-R1-70B} performs most reliably, achieving 93.8\% overall coverage with compact sets, and generalizes well across specialties and question types. \texttt{Llama3-OpenBioLLM-70B} also reaches high coverage (90.1\%) but with larger sets compared to \texttt{DeepSeek-R1-70B}, and it fails to meet the target in several categories, particularly Infectious Diseases and Definition questions.

In contrast, \texttt{Llama-3.1-70B} falls short of the coverage target (85.5\%) with highly restrictive predictions, indicating overconfidence and poor generalization from calibration to test data. For this model, no specialty or question type meets the 90\% threshold, suggesting that its strong score-based UQ performance does not extend to set-based methods. These findings underscore that accuracy alone does not suffice. Robust uncertainty estimation depends on careful calibration and generalization across specialties, question types, and uncertainty paradigms, particularly in light of the misalignment observed between score- and set-based methods.

%% file: parts/6-conclusions.tex
\section{Discussion and Conclusion}
\label{sec:conclusion}

We investigate uncertainty estimation for LLMs in clinical QA. By jointly assessing calibration and discrimination spanning diverse models and across clinical specialties and question types, we address key limitations of prior work. Our findings show that model performance varies not only in accuracy but also in uncertainty quality, with substantial differences across clinical contexts. Reasoning-oriented LLMs often perform well, but no method succeeds consistently across settings; evaluation and deployment must be context-specific. 

In conclusion, our work raises a crucial cautionary note for the deployment of LLMs in clinical NLP, while also offering a constructive path forward. Our analysis demonstrates that a model's ability to signal uncertainty is not a monolithic capability but one that varies significantly with the clinical context. To support more nuanced evaluation, we propose a lightweight and interpretable UQ case study based on behavioral cues from reasoning traces, which approximates the performance of more resource-intensive sampling-based methods. The primary lesson for NLP practitioners is that for high-stakes domains, `one-size-fits-all' evaluations are insufficient and potentially misleading. The true measure of an LLM's reliability lies in its performance on the specific tasks and in the precise domains where it will be used. Future work should involve clinicians in assessing questions flagged as uncertain, to determine which signals best indicate when additional human review is warranted.